\documentclass[letterpaper, 10 pt, conference]{ieeeconf}  

\IEEEoverridecommandlockouts                              

\usepackage{graphicx} 
\usepackage{subcaption} 
\usepackage{todonotes}
\usepackage{hyperref}

\usepackage{svg}

\title{\LARGE \bf
From the Lab to the Theater: An Unconventional Field Robotics Journey
}

\author{Ali Imran$^{1}$, Vivek Shankar Varadharajan$^{2}$, Rafael Gomes Braga$^{1}$, Yann Bouteiller$^{2}$, \\ Abdalwhab Bakheet Mohamed Abdalwhab$^{1}$, Matthis Di-Giacomo$^{1}$, Alexandra Mercader$^{1}$, \\ Giovanni Beltrame$^{2}$ and David St-Onge$^{1}$. 
\thanks{*This work was supported by FRQ PRISME-ART program.}
\thanks{$^{1}$ are with the Lab of INIT Robots, Mechanical Engineering, Ecole de technologie supérieure, 1100 Notre-Dame W., Canada
        {\tt\small name.surname@etsmtl.ca}}%
\thanks{$^{2}$ are with MISTLab, Computer and Software Engineering, Polytechnique Montréal, Canada
        {\tt\small name.surname@polymtl.ca}}%
}

\usepackage{fancyhdr}
\fancypagestyle{withfooter}{
  
  \fancyfoot[C]{\footnotesize Accepted to the IEEE ICRA Workshop on Field Robotics 2024}
}

\begin{document}

\maketitle
\thispagestyle{withfooter}
\pagestyle{withfooter}

\begin{abstract}

Artistic performances involving robotic systems present unique technical challenges akin to those encountered in other field deployments. In this paper, we delve into the orchestration of robotic artistic performances, focusing on the complexities inherent in communication protocols and localization methods. Through our case studies and experimental insights, we demonstrate the breadth of technical requirements for this type of deployment, and, most importantly, the significant contributions of working closely with non-experts.

\end{abstract}

\section{Introduction}

The intersection of art and technology has given rise to a burgeoning field of robotic artistic performances (such as ~\cite{Vorn2016,meritt2020,cuan2020}). From synchronized choreographies to interactive installations, these performances push the boundaries of traditional artistic expression, inviting audiences to engage with the fusion of human creativity and robotic technology. However, behind the curtain of artistic innovation lies a host of technical challenges that rival those encountered in more conventional field deployments of robotic systems.

Artistic deployment of robotics shares many common aspects with the more conventional fields of robotics. First and foremost, accurate localization methods are crucial in any deployment of robots outside a controlled lab environment. The operating environment could be outdoors with access to GPS, or it might be a GNSS-denied setting \cite{braga2017coordinated}, such as in mining applications, cave exploration, or industrial environments.

Over the last decade, major field deployment targeted robot swarms, such as the DARPA OFFSET~\cite{rouvcek2020darpa, ju2022review}. In these settings, communication is one of the most critical aspects for coordination and information sharing \cite{mourikis2006performance, cortes2017coordinated}.

In this paper, we explore the technical challenges of orchestrating robotic artistic performances based on the use case of an artistic project called DESSAIM\footnote{\href{https://initrobots.ca/en/researches/ecriture-choregraphique-des-esssaims}{DESSAIM web page}}. DESSAIM stems from an interdisciplinary research project combining engineering and performing arts, exploring the expressiveness of three types of robotic swarms. Recycled materials transform
the robots' appearance and the environment in which they operate. From these imaginary territories, the poetry of material and form in motion emerges.

As this project calls for the deployment of self-organised multi-robot systems in close proximity to the audience, our core technical challenges were posed by communication protocols and localization methods in dynamic performance spaces as well as safety. While the primary focus of robotic performances may be on aesthetic expression and emotive storytelling, the underlying technical infrastructure is critical to the seamless execution of representations.

This paper first describes the approach used to design swarm behaviors, then the robotic systems deployed in this field experiment, and finally the technical challenges and their solutions in terms of communication and localization.


\section{Behavior composition}

The team of artists delved into swarm behaviors and their inspiration from nature to create the performances. We explored together the well-known aggregation, diffusion, flocking \cite{schranz2020swarm}, Lennard-Jones~\cite{rosbuzz2020} and pursuit~\cite{pursuit2018} behaviors. In the end, our team created a library of twelve emergent behaviors, to be applied to three swarms of robots at different scales: small tabletop, medium flying robots, and wheeled bases with manipulators. They each call upon swarm behavior primitives, but with variations in dynamic states and combinations that extend their expressivity. All these behaviors were programmed in Buzz~\cite{pinciroli2016buzz}, a domain-specific language for robot swarms, and validated in a low-fidelity fast simulator, ARGoS3~\cite{PinciroliSI2012}. The use of these two tools was instrumental in sharing knowledge on swarm intelligence and iterating on their behavior compositions.

\begin{figure}[ht]
  \centering
  \includegraphics[width=0.44\columnwidth]{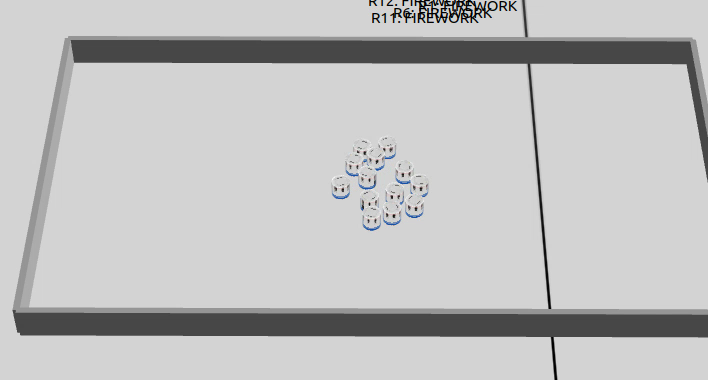}\quad
  \includegraphics[width=0.44\columnwidth]{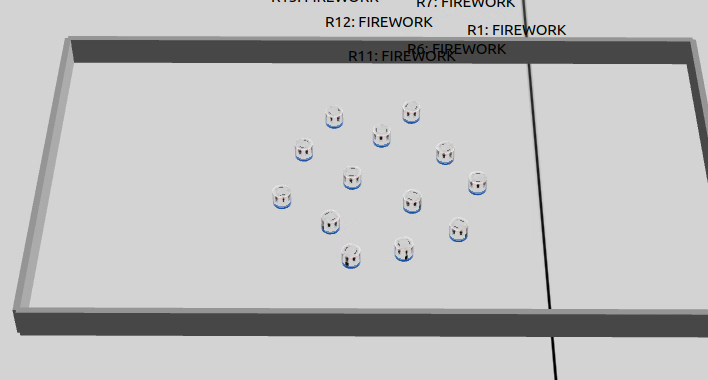}
  \includegraphics[width=0.44\columnwidth]{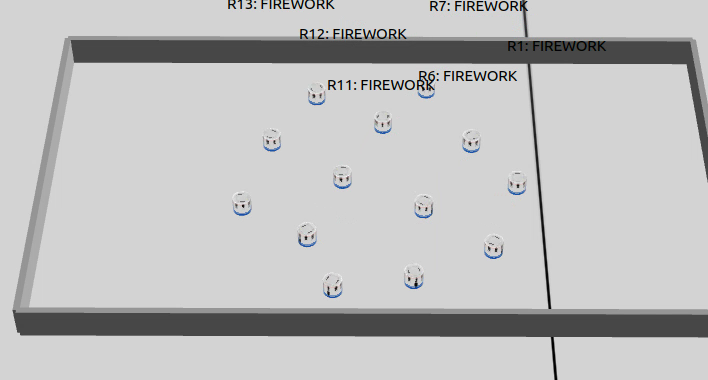}\quad
  \includegraphics[width=0.44\columnwidth]{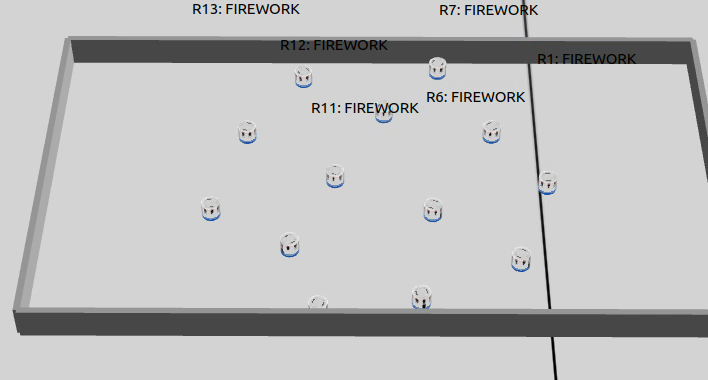}
  \caption{Behavior composition: A ``firework'' composed of 4 behaviors in ARGoS}
  \label{Firework}
\end{figure}

As an example, Fig.~\ref{Firework} presents the firework behavior where all robots are attracted towards a point and then spread simultaneously in opposite directions like an expanding circle. Velocities and timing in both directions are different and finely tuned to create the desired effect.

\begin{figure*}[htp]
    \centering
    \subcaptionbox{Tabletop swarm: Sushis\label{fig:sushis}}{%
        \includegraphics[height=2.25in]{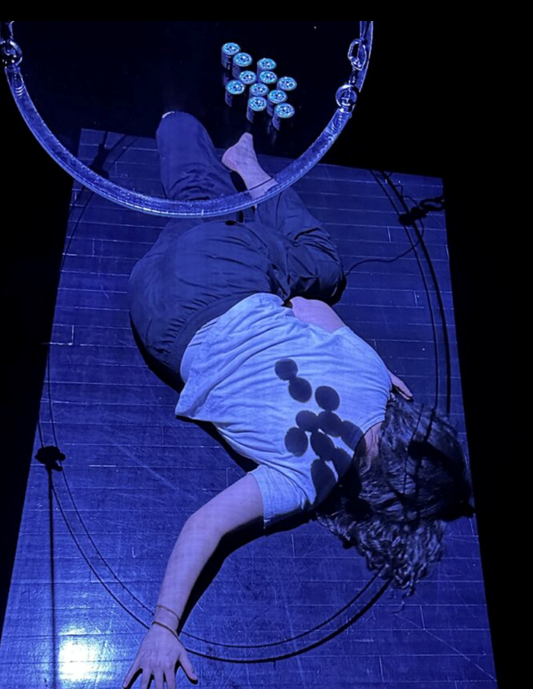}%
    }\quad
    \subcaptionbox{Flying swarm: CrazyCognies\label{fig:cc}}{%
        \includegraphics[height=2.25in]{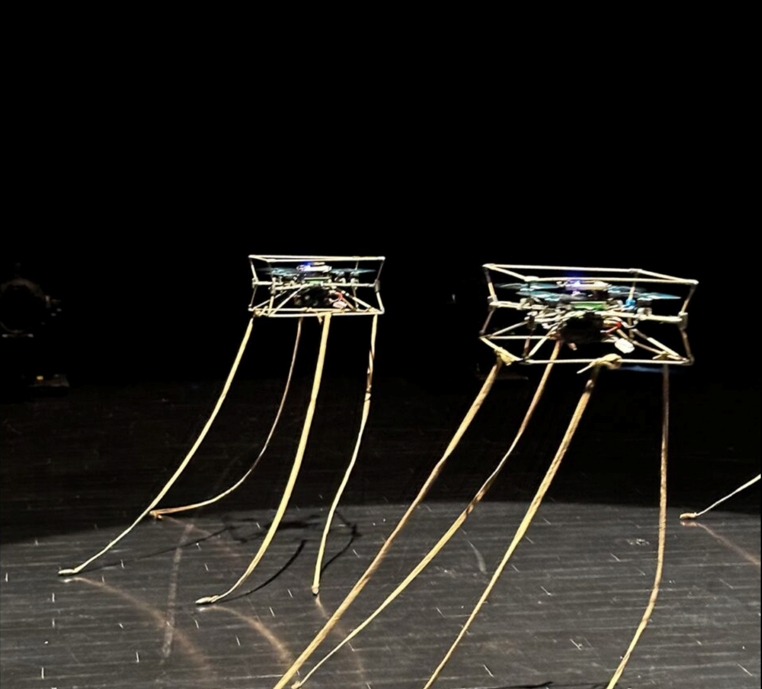}%
    }\quad
    \subcaptionbox{Human-scale swarm: Doodies\label{fig:Doodies}}{%
        \includegraphics[height=2.25in]{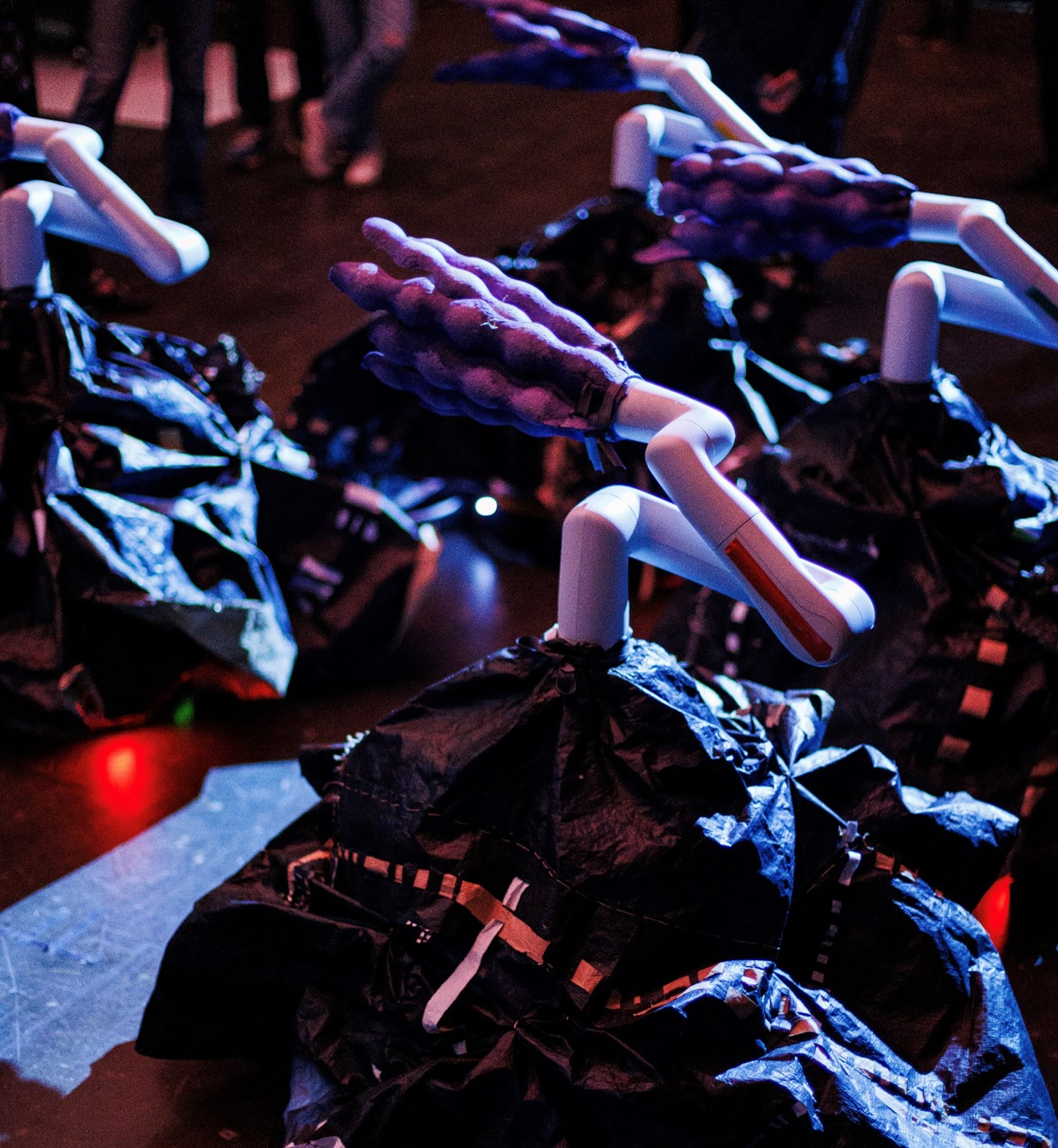}%
    }
    \caption{The three DESSAIM robotic swarms seen in their performing settings.\label{fig:swarms}}
\end{figure*}

The three platforms introduced in the following section run the exact same Buzz scripts, robot-agnostic, in a Python wrapper adapted to their hardware called (Py)Buzz. (Py)Buzz is our swarm controller developed for artistic performances; it implements the Buzz Virtual Machine (BVM)~\cite{pinciroli2016buzz} to run collective behaviors through Buzz scripts. 

 


\section{DESSAIM swarms}

As shown in Fig.~\ref{fig:swarms}, three swarms of various sizes and dynamics performed in DESSAIM: tiny tabletop Sushis, flying caged CrazyCognies, and human-size mobile manipulators, the Doodies.  

\subsection{Tabletop swarm: Sushis}
Our Sushis are small tabletop robots derived from Stanford's Zooids \cite{le2016zooids}. 

\begin{figure}[!h] 
\centering
\includegraphics[width=0.5\columnwidth]{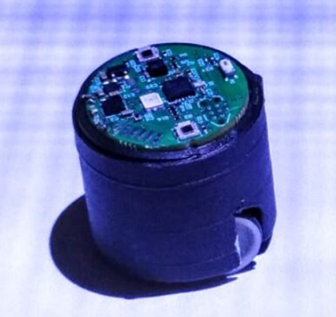}
\caption{Sushi with its top circuit including RF communication chip and photodiode for localization.}
\label{ZooidComponents}
\end{figure}

A 48MHz ARM microcontroller (STM32F051C8) manages onboard control and communication using a 2.4GHz radio chip (NRF24L01+). For localization, two photodiodes are placed on top to detect and decode gray-coded patterns projected from the top by a 3000Hz projector (DLP LightCrafter 4500). Two small DC motors actuate their motion in differential drive mode. Their small 100 mAh LiPo battery can withstand at least one hour of continuous performance.


Our revision of the original design includes a larger enclosure to facilitate their assembly and maintenance as well as to accommodate a larger gearbox for the drive motors. The expressivity of their motion is highly coupled to their dynamics (velocity and acceleration), which required a larger band of possible velocities, provided with the new gearbox. In our setup, their control is now tied to (Py)Buzz.

\subsection{Flying swarm: CrazyCognies}
\begin{figure}[!h] 
\centering
\includegraphics[width=0.8\columnwidth]{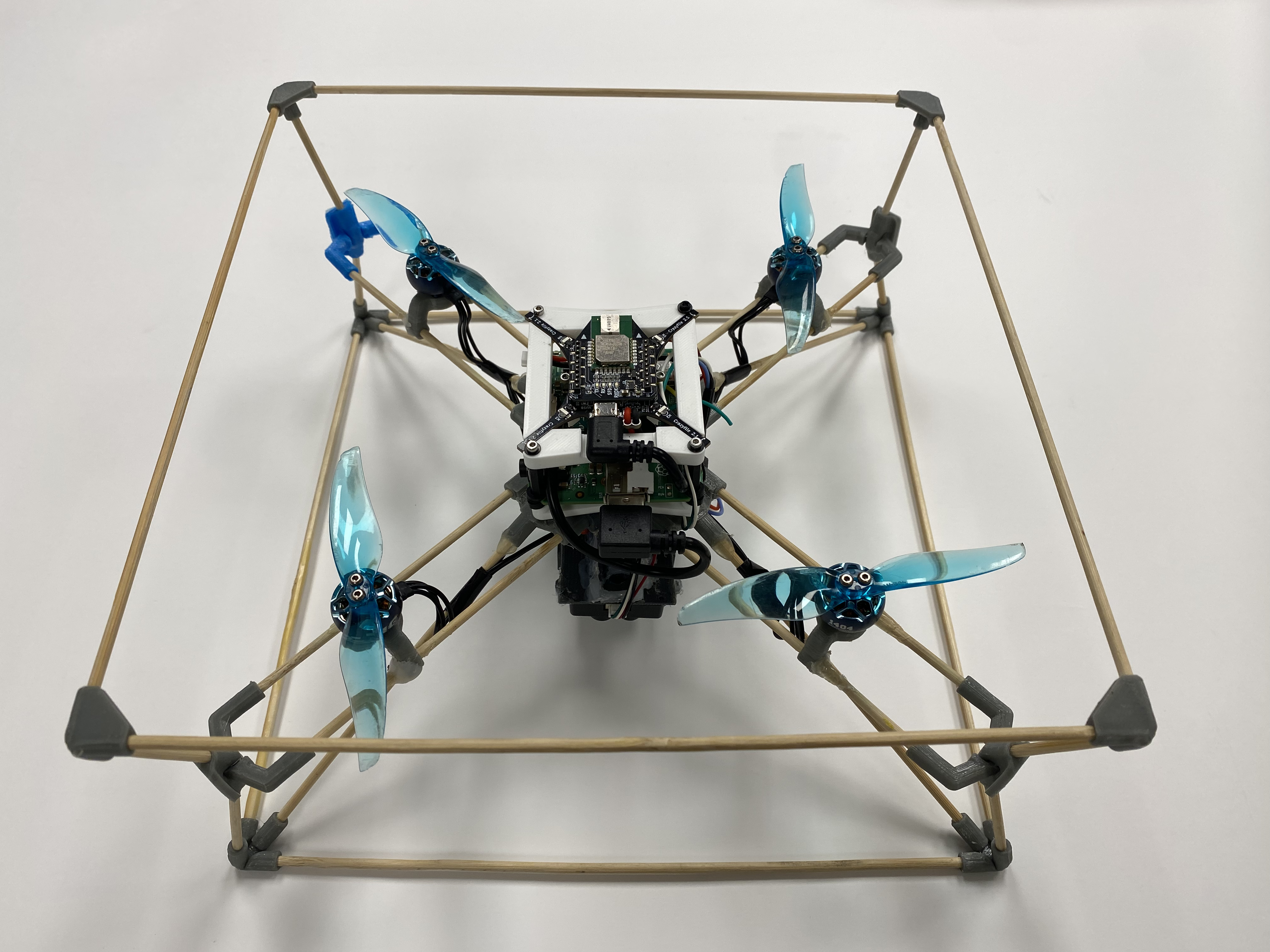}
\caption{CrazyCognie with its Crazyflie on top, a Raspberry Pi just underneath, and the iNav flight controller below, just above the battery pack.}
\label{CogniComponents}
\end{figure}

Our Crazycognies are hybrid indoor aerial robots made from the combination of Cogniflies~\cite{de2022being}, an open-sourced collision-resistant quadcopter, and the Crazyflie quadcopter commercialized by BitCraze. We developed these platforms for safe deployments in tight and cluttered environments - a requirement of indoor artistic performances. The Crazyflies were not powerful enough for our needs (flight time and payload), whereas the Cogniflies were not ready for advanced indoor positioning. 
The resulting system shown in Fig.~\ref{CogniComponents} weighs 310 grams and can fly reliably for about 7 minutes.

\begin{figure}[ht!] 
\centering
\includegraphics[width=0.4\textwidth]{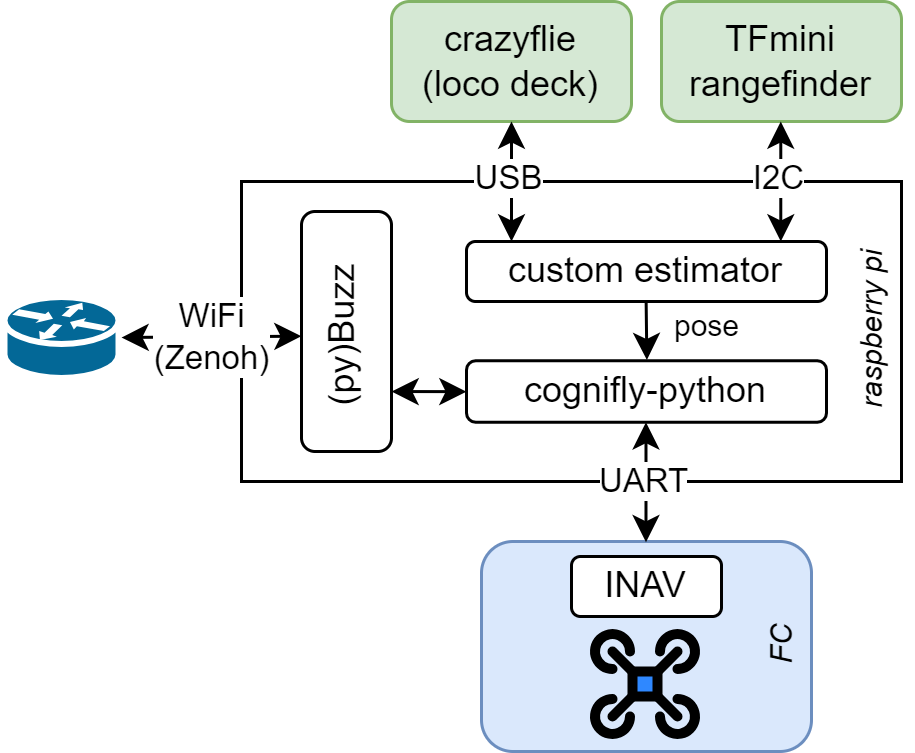}
\caption{Software schematic of a CrazyCogni}
\label{fig:CrazyCogni}
\end{figure}

Figure~\ref{fig:CrazyCogni} shows a schematic of the software components deployed on the Crazycognies. The driver software (cognifly-python\footnote{\href{https://github.com/thecognifly/cognifly-python}{GitHub code repository for Cognifly-Python}}) communicates with the Flight Controller (FC) through UART, and that software is designed to enable the deployment of custom pose estimation. The custom estimator communicates with the Crazyflie through USB for horizontal pose estimates (X, Y) and sends it to the FC to achieve stable flight in the theater room. 
As for the altitude, the highly unreliable nature of the altitude estimates from UWB required us to use TFmini lidar measurements instead.

\subsection{Human-scale swarm: Doodies}
\begin{figure}[ht!] 
\centering
\includegraphics[width=0.5\columnwidth]{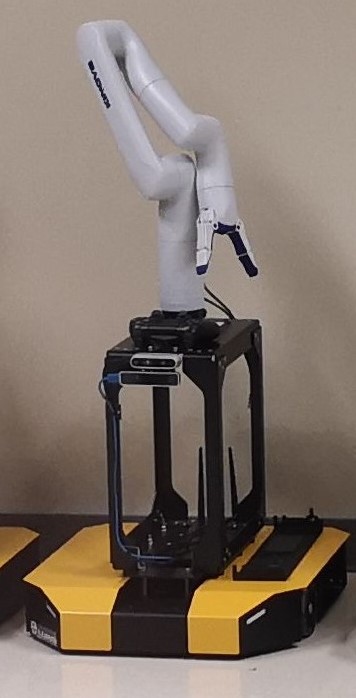}
\caption{Doody}
\label{fig:dingonu}
\end{figure}

Our swarm of Doodies consists of a fleet of six Clearpath Dingo \cite{clearpath_dingo_2024} differential drive robots, each equipped with a Kinova Gen3 Lite \cite{kinova_gen3_lite_2024} manipulator. Similar to the Crazycognies, the Doodies are also equipped with a loco Crazyflie deck for position estimation. They run (Py)Buzz, our custom Kalman filter, and all hardware drivers on an onboard Nvidia Xavier. For safety, the platform uses small 12V Lead-acid batteries, one for the base and another for the arm, granting autonomy for continuous motion of more than 3 hours. As shown in Fig.~\ref{fig:dingonu}, the custom turret brings the whole robotic platform to a human scale and includes an emergency stop in the back to increase safety.

\begin{figure}[ht!] 
\centering
\includegraphics[width=0.4\textwidth]{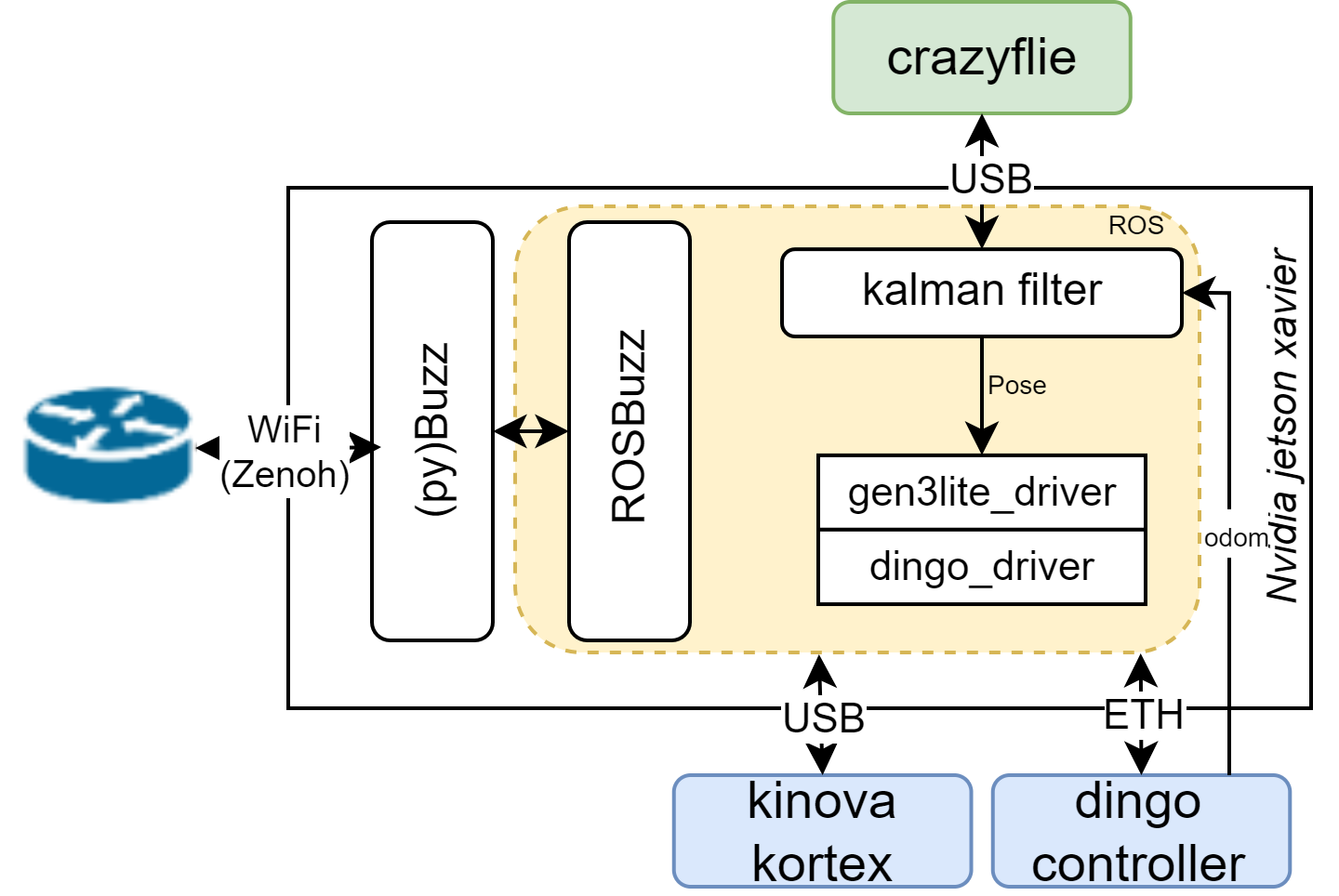}
\caption{Software Schematic of a Doody}
\label{DoodySoftware}
\end{figure}

Figure~\ref{DoodySoftware} shows the software architecture on each robot built over modified versions of the drivers for the Clearpath Dingo base, Kinova Kortex for the robotic arm, Crazyflie \cite{bitcraze_crazyflie_2024} firmware and node to leverage UWB localization, and drivers for the cameras to provide interaction capabilities. On the Doodys, (Py)Buzz is bridged with the ROS ecosystem.
Localization of the Doodies is achieved with a Kalman filter fusing the wheel odometry, the IMU measurements, and the 2D pose estimates from the UWB setup. The resulting pose is sent to (Py)Buzz for the swarm-level collective behaviors.


\section{Artistic performance challenges}
The three swarms undergo five iterative performances in front of an audience of 30 people throughout the creation process. Each performance concludes with a mediation session with the public to describe the robot functionalities and gather their feedback and perception. The creation process culminates with the presentation of five identical performances in front of audiences of 45 people as part of a local festival in Montréal, Canada.

\subsection{High-frequency critical communication for coordination}
Coordination among robots to synchronize their states and achieve collective behaviors relies heavily on inter-robot communication.

Reliable communication in the application layer of the ISO/OSI model can be achieved using TCP/IP protocols; some existing packages like nimbro\_network~\cite{github_nimbro} enable it with ROS. However, the configuration task complexity increases rapidly with the growing number of robots and is challenging for large swarms. We thus selected Zenoh~\cite{zenoh}, a publish/subscribe/query protocol that operates with a set of abstractions in the 4 and 5 layers of the OSI model.

\begin{figure}[ht]
  \centering
  \includegraphics[width=0.95\columnwidth]{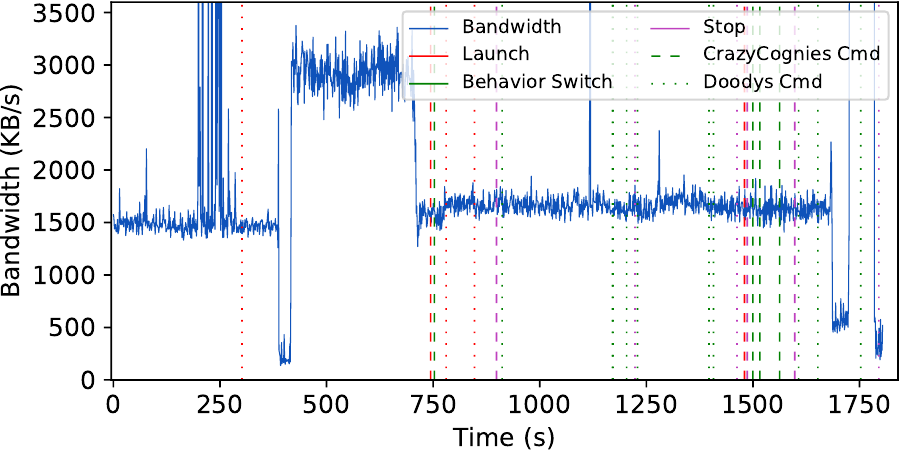}
  \caption{The figure reports the bandwidth used to operate and maintain a swarm of five Doodys, five CrazyCognies, a mobile marker node (used as an attractor/repulsor point for behaviors), and 2 ground stations. The vertical dotted lines (Doodys) and dashed lines (CrazyCognies) correspond to specific events: a launch command (red), behavioral state switch command (green), and behavioral stop command (purple).}
  \label{fig:bandwidth}
\end{figure}

In the application layer, we implement publishers and subscribers for each robot's gossip messages through a single topic, enabling us to add and remove robots seamlessly. Zenoh requires only a small communication overhead of 5 bytes per packet. Figure~\ref{fig:bandwidth} shows the recorded communication bandwidth during one of the artistic performances. The bandwidth spikes to over 4 MB/s when connecting to the Doodys and launching the software around 250s, similarly around 500s when launching the Crazycognies. During the normal operation of both swarms, the bandwidth stays below 2 MB/s since each node in the swarm sends gossip below 250 bytes. When both the swarms are shut down, the bandwidth drops to below 0.5 MB/s. The communication setup introduced in this work was proven to be stable to swarms with up to 13 nodes and exhibits graceful communication scalability.

\subsection{Minimal infrastructure global positioning}

Relative localization is vital for robots to achieve collective behaviors. Range and bearing information between neighboring robots is often enough for swarm behavior deployment, but most technologies (IR, vision, etc.) require line-of-sight; a complex thing to achieve with tens of people walking among the robots. Furthermore, in a public performance, the robots always need to undergo safe motion with respect to the theater equipment (speakers, lightning, curtains) and the people. However, deploying a vision-based motion capture system is not an option: the artistic shows change venues often. Each performance has a varying stage lighting design. Moreover, the rooms are rather large to cover (for instance, 6m x 12m).
We opt for a UWB-based strategy, using preconfigured anchors in the environment, to derive a rough global estimation. The anchors were configured in TDOA3 \cite{zhao2021learning} mode to ping the robots' UWB tags. The resulting estimates obtained from the UWB undergo another layer of filtering, which is particularly important for robots that are more sensitive to noise - such as the Crazycognis. 

\begin{figure}[ht]
  \centering
  \includegraphics[width=0.95\columnwidth]{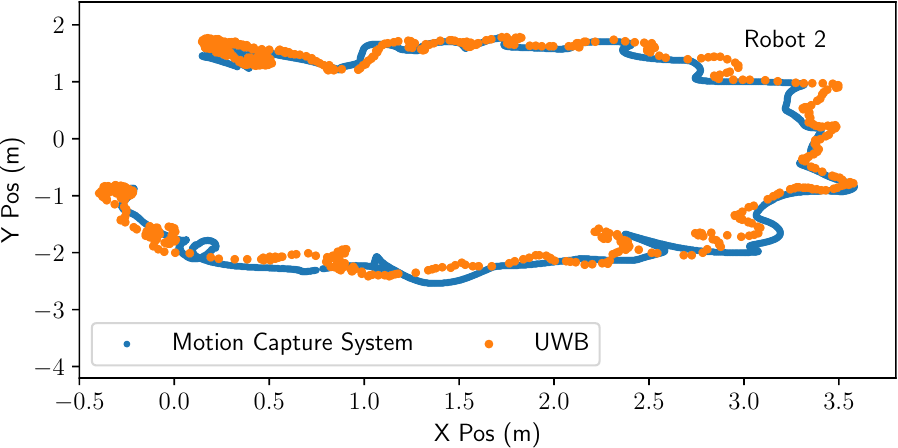}
  \caption{Comparison of our UWB pose estimation (orange) and ground truth from an Optitrack motion capture (blue) system during a pursuit behavior executed by 5 CrazyCognies.}
  \label{fig:optivsuwb}
\end{figure}

Figure~\ref{fig:optivsuwb} compares the pose estimates of a robot when 5 Cogniflies display a pursuit behavior. The pose estimates were obtained using UWB and a motion capture system installed at the venue. The UWB-based pose estimates show errors in the range of tens of centimeters, with occasional fluctuations caused by the abrupt drone movements. The accuracy of the UWB-based systems is significantly influenced by factors such as the initial calibration, the number of anchors, and their configuration (gain, bandwidth, etc.). To enhance this, our team has developed advanced calibration strategies, inspired by land surveying techniques, enabling us to install our affordable UWB system (\textless\$1500) in less than 4 hours.




\section{CONCLUSIONS}

We describe the three robotic swarms performing in the DESSAIM artistic representations. Each of the swarms was running the exact same behavioral script, as designed (composed) by the artists and the engineers. About 400 people experienced this intimate and immersive robotic performance, with the audience freely moving among the performing robots. To achieve robustness (capable of repeated performance) and safety, this paper highlighted two key aspects: the Zenoh communication layer integration and the UWB localization. We believe this report serves as a compelling demonstration of the value of artistic manifestations as challenging field robotic deployments.

\addtolength{\textheight}{-12cm}   



\section*{ACKNOWLEDGMENTS}

The authors would like to acknowledge the creative and empathetic contribution of the talentful artists of DESSAIM: Audrey Rochette, Hélène Duval, Marcelle Hudon, Magali Babin and Danielle Lecourtois. Several other engineering students support this team effort: Jean Mazerolle, Corentin Boucher, Hao-Fang Cheng, Clement Gassmann-Prince, Guillaume Ricard and Radhouen Khmiri.


\bibliographystyle{IEEEtran}
\bibliography{references}

\end{document}